\documentclass{article}

\usepackage{microtype}
\usepackage{graphicx}
\usepackage{subfigure}
\usepackage{booktabs}

\usepackage{listings}
\usepackage{multirow}
\usepackage{multicol}
\usepackage{pgfplots}
\usepackage[table]{xcolor}
\definecolor{highlight}{HTML}{A7D7D7}
\definecolor{plotcolour1}{HTML}{8dd3c7}
\definecolor{plotcolour2}{HTML}{fdb462}
\definecolor{plotcolour3}{HTML}{bebada}
\definecolor{plotcolour4}{HTML}{fb8072}
\definecolor{plotcolour5}{HTML}{80b1d3}
\definecolor{plotcolour6}{HTML}{ffd92f}

\usepackage{hyperref}

\usepackage[accepted]{icml2026}

\usepackage{amsmath}
\usepackage{amssymb}
\usepackage{mathtools}
\usepackage{amsthm}

\usepackage[capitalize,noabbrev]{cleveref}

\theoremstyle{plain}

\theoremstyle{definition}

\theoremstyle{remark}

\usepackage[textsize=tiny]{todonotes}

\icmltitlerunning{Reasoning Before Translation: Enhancing Legal Machine Translation with Structured Reasoning}

\begin{document}

\twocolumn[
\icmltitle{Reasoning Before Translation: Enhancing Legal Machine Translation with Structured Reasoning}

\icmlsetsymbol{equal}{*}

\begin{icmlauthorlist}
\icmlauthor{Aixiu An}{equal,hefr}
\icmlauthor{Michael Jungo}{equal,unifr}
\icmlauthor{Eloi Eynard}{neuron}
\icmlauthor{Mark Drenhaus}{Humanist}
\icmlauthor{Andreas Fischer}{unifr}
\icmlauthor{Jean Hennebert}{hefr}
\icmlauthor{Sébastien Rumley}{hefr}
\end{icmlauthorlist}

\icmlaffiliation{hefr}{iCoSys, University of Applied Sciences and Arts Western Switzerland, Fribourg, Switzerland}
\icmlaffiliation{unifr}{AIBEX, University of Fribourg, Fribourg, Switzerland}
\icmlaffiliation{neuron}{Neur.on, Fribourg, Switzerland}
\icmlaffiliation{Humanist}{Human-IST, University of Fribourg, Fribourg, Switzerland}

\icmlcorrespondingauthor{Aixiu An}{aixiu.an@hefr.ch}
\icmlcorrespondingauthor{Michael Jungo}{michael.jungo@unifr.ch}

\icmlkeywords{Machine Learning, ICML, AI4Law, Legal Translation, RL, SFT, Reasoning}

\vskip 0.3in
]

\printAffiliationsAndNotice{\icmlEqualContribution}

\begin{abstract}
Neural machine translation (NMT) in the legal domain is a linguistically and conceptually  demanding task, primarily due to the complexity of legal language and the high level of precision it requires. The recent emergence of reasoning-capable language models opens new possibilities for tackling such challenges. 
They add to a set of other previously proposed techniques to enhance the translation quality, which includes supervised fine-tuning and reinforcement learning. 

In this work, we perform a comparison between these various approaches. More particularly, we evaluate small language models such as \texttt{Qwen3.5 4B}, \texttt{Qwen3.5 9B}, and \texttt{Gemma 3 12B} enhanced with various \emph{re-training paradigms} and compare their performances against frontier reasoning models. We focus on the Swiss legal system, which---with its unique  multilingual statutes---offers a particularly challenging testbed for reasoning-augmented models. Our results show that the quality of small ``base'' models can be greatly enhanced, and that reinforcement learning with verifiable rewards \emph{can be applied to NMT in the legal domain and surpasses the translation quality of supervised fine-tuning}. The performance of enhanced small models is close to the one of state-of-the-art reasoning models yet remains inferior.  
We also note that \emph{re-training paradigms} yield diminishing returns as model size increase. 
The code and models are publicly available at \href{https://github.com/aixiuxiuxiu/Legal-MT-SFT-RL}{https://github.com/aixiuxiuxiu/Legal-MT-SFT-RL}.

\end{abstract}

\section{Introduction}

Neural Machine Translation (NMT) has progressed rapidly alongside the development of large language models (LLMs), achieving substantial improvements in translation quality across many domains and language pairs \citep{kocmi2023gemba,finkelstein2026translategemma}. These advancements are particularly relevant in the context of multilingual legal translation, where precision, consistency, and preservation of legal meaning are critical. In Switzerland, the coexistence of German, French, Italian and Romansh as official languages, together with the multilingual nature of the Swiss legal system, creates a particularly challenging environment for NMT systems \citep{martinez2020customized, canavese2024translators, niklaus2025swiltra}.

Beyond multilingualism, legal texts themselves introduce additional challenges. They are characterized by specialized terminology, complex syntactic structures, and the formal register of \textit{legalese}—a style of legal writing often marked by highly technical vocabulary, long and syntactically complex sentences, and precise but rigid formulations \citep{mattila2016comparative, tiersma2008nature, martinez2023even}. These linguistic features—combined with the need to preserve legal accuracy, nuance, and internal consistency—pose significant difficulties not only for human translators, but also for machine translation systems \citep[e.g.,][]{koehn2017six, wiesmann2019machine}, and continue to push the limits of even the most advanced NMT models.

Recent work has demonstrated that LLMs, such as GPT, exhibit strong general-purpose translation capabilities with zero- and few-shot prompting \citep{hendy2023good, jiao2023chatgpt}. Building on these capabilities, some approaches have attempted to adapt small language models (SLMs) to the legal domain by fine-tuning them on domain-specific corpora \citep{niklaus2025swiltra}. Such approaches are particularly relevant in legal and institutional contexts, where computational efficiency, deployment cost, data privacy, and on-premise inference constraints may limit the practical use of large proprietary models. While these methods yield improved performance over non-fine-tuned models, they often fall short of state-of-the-art LLMs in capturing the full linguistic complexity and precision required for high-quality legal translation.

More recently, advances in NMT have been significantly influenced by the emergence of Large Reasoning Models (LRMs), such as the GPT-O series and DeepSeek-R1~\cite{deepseek-r1}, which have opened new opportunities for addressing complex translation tasks~\cite{liu2025new}. These models have demonstrated impressive performance on reasoning-intensive tasks such as mathematics~\cite{math-reasoning-progress, math-reasoning-beyond-accuracy} and programming~\cite{competitive-programming, code-reasoning-hypothesis-decomposition}, highlighting their capacity to reflect, reason, and iteratively refine their outputs. In the context of translation, LRMs extend their reasoning capabilities by treating translation as a dynamic inference process—enabling more accurate handling of context, cultural nuances, and syntactic complexity. For instance, their self-reflective mechanisms can be leveraged for error detection and correction during inference, allowing them to better handle ambiguous and linguistically demanding translation scenarios \cite{wang2024drt,chen2025evaluating,wu-etal-2025-please}.

In light of these developments, this paper explores how reasoning can be leveraged to improve translation quality, with a particular focus on Swiss legal translation. We investigate both the impact of different \textbf{training paradigms} and the role of \textbf{model scale} in reasoning-based legal translation. In particular, we compare supervised fine-tuning (SFT) and reinforcement learning (RL) approaches across models of different sizes, ranging from 4B to 12B parameters, to better understand how reasoning capabilities interact with model capacity.

In the SFT setting, we investigate two variants: \textbf{Reasoning SFT}, where models are fine-tuned on translation pairs accompanied by intermediate reasoning steps; and \textbf{Simple SFT}, where models are trained solely on direct sentence-level translation pairs. We then compare these approaches with an RL setup based on Group Relative Policy Optimization (GRPO)~\cite{shao2024deepseekmath} to evaluate the effectiveness of reasoning-augmented training strategies for legal translation across different model scales.

Specifically, we make the following three contributions:

\begin{enumerate}
\item We conduct a systematic and detailed comparison of three approaches for integrating reasoning into translation across different model families: (1) prompting large reasoning models, (2) Fine-tuning of small language models using intermediate reasoning steps, and (3) applying RL techniques to optimize model's translation.

    \item We release a new dataset of 40\,000 legal translation pairs enriched with automatically generated intermediate reasoning steps, produced using the DeepSeek-R1 model. This dataset aims to support future research on reasoning-augmented translation in legal domains.
    
    \item We propose a GRPO training recipe, accompanied by a simple yet effective reward formulation tailored to legal translation tasks, which can serve as a blueprint for future work on domain-specific reinforcement learning in NMT.

\end{enumerate}

Our main findings are as follows:  
(a) While both SFT and RL improve translation quality, these effects are particularly pronounced for smaller models;  
(b) RL yields more substantial performance gains than SFT;  
(c) Reasoning-augmented SFT does not consistently outperform simple SFT;  
(d) Model size plays an important role, and among the evaluated smaller-scale models, GRPO-trained Gemma 3 12B achieves the best performance for Swiss legal translation. \footnote{The code is available in the GitHub repository \url{https://github.com/aixiuxiuxiu/Legal-MT-SFT-RL}}

\section{Related Work: Reasoning in Neural Machine Translation}

Recent studies have begun to leverage reasoning capabilities of language models to enhance NMT, with approaches ranging from multi-agent systems to reinforcement learning. For example, \citet{wang2024drt} propose a multi-agent framework for translating figurative language, where a translator agent iteratively improves translations based on feedback from advisor and evaluator agents. Similarly, \citet{wu-etal-2025-please} investigate reasoning-based prompting for MT and find that simple self-refinement strategies, such as asking the model to “translate again,” can outperform explicit chain-of-thought decomposition.

Reinforcement learning  for deep reasoning in MT is a relatively new area, but emerging studies reflect growing interest in its potential. A notable example is R1-T1 \citep{he2025r1}, which uses COMET scores \citep{rei2020comet} as reward signals and trains models via a modified REINFORCE++ algorithm \citep{hu2025reinforce++}. Building on this work, \citet{wang2025deep} enhance reasoning-capable LLMs using RL, adopting the Group Relative Policy Optimization (GRPO)~\cite{shao2024deepseekmath} as the training foundation. Rather than relying on predefined heuristic metrics, they employ an advanced LLM as a reward model, which evaluates both the translation output and the reasoning process using carefully designed prompt-based scoring criteria. More recently, \citet{finkelstein2026translategemma} introduced the TranslateGemma model, which combines supervised fine-tuning with reinforcement learning, where the latter stage optimizes translation quality using an ensemble of reward models, including MetricX-QE \cite{juraska-etal-2024-metricx} and AutoMQM \cite{fernandes2023devil}.

\section{Experiments Setup}

\subsection{Datasets} \label{sec::dataset}

Our experiments build upon the \texttt{SwiLTra-Bench} dataset~\citep{niklaus2025swiltra}, which is currently the largest publicly available dataset for legal translation in Switzerland. The dataset spans five languages: the four official Swiss languages (German, French, Italian, and Romansh) as well as English. For this study, we focus on the \texttt{SwissLawTranslations} (CH-Law-Trans) subset, which includes approximately 631k aligned translation pairs for training and 22.7k sentence pairs for evaluation.

\paragraph{Training Dataset} To construct a training dataset suitable for Reasoning SFT, we use the \texttt{DeepSeek-R1} model to generate intermediate reasoning steps for each translation example. Specifically, we select the first 40\,000 sentence pairs from the training split of \texttt{SwissLegalTranslations} and prompt DeepSeek-R1 to produce both the reasoning steps and the final translation using prompt format:

\begin{quote}
\textit{
Translate the following sentence to \textbf{target language} while respecting Swiss legal parlance.} \\[1ex]
\textit{
Give the final translation starting with ``Final translation:''} \\[1ex]
\textit{
Here's the sentence to translate:} \\[1ex]
\textit{
Source sentence: \textbf{source sentence}}
\end{quote}

To ensure that the reasoning steps meaningfully contribute to correct translations, we evaluate the final outputs using the ChrF~\cite{popovic2015chrf} score, a character n-gram overlap metric commonly used for machine translation evaluation \footnote{We use the ChrF score here because of its effectiveness in capturing lexical and terminological accuracy, which is particularly important in the legal domain.}. We retain only samples with a ChrF score above the median value (64.19), thereby filtering out lower-quality translations while preserving approximately half of the generated data. This filtering step helps isolate reasoning sequences that are more likely to support high-quality translation. The resulting dataset contains 19\,979 sentence pairs, which are split into 90\% for training and 10\% for validation. The distribution of language pairs in this filtered dataset is illustrated in \autoref{tab:lang-pair-counts}.

\paragraph{Evaluation Dataset}  
For the evaluation across the three experiments described in the following section, we use the 22.7k sentence pairs from the test split of \texttt{SwissLegalTranslations}. To ensure comparability, we exclude Romansh from the dataset, as it is not supported by the translation models under consideration. After filtering, the final evaluation dataset consists of 18.1k sentence pairs.

\subsection{Evaluation Metrics}

For evaluation, we use a combination of commonly used machine translation evaluation metrics, including both lexical and model-based approaches: ChrF~\cite{popovic2015chrf}, METEOR~\cite{banerjee2005meteor}, COMET~\cite{rei2020comet}, and MetricX~\cite{juraska-etal-2024-metricx}. In contrast to ChrF, which measures character n-gram similarity between the generated translation and the reference, METEOR extends lexical matching by incorporating stemming and synonymy, allowing for a more flexible comparison between translations and references. COMET is a neural evaluation metric that leverages pretrained multilingual language models to assess semantic adequacy and fluency by comparing the source, hypothesis, and reference translation. MetricX is a large language model-based evaluation metric designed to provide fine-grained assessments of translation quality, with a stronger focus on semantic correctness and factual consistency.

While GEMBA-MQM~\cite{kocmi2023gemba} has demonstrated strong correlation with human judgments, we did not prioritize it because it relies on GPT-based evaluation and would incur a high estimated evaluation cost exceeding \$3\,000. In addition to automatic metrics, we also perform human evaluation on a representative sample of translations to further assess translation quality and validate the automatic evaluation results.

\section{Experiments}
\subsection{Experiment I: Reasoning via Prompting}

To assess the capabilities of current state-of-the-art large reasoning models in legal translation, we evaluate a diverse set of model families, including \texttt{DeepSeek-R1}, OpenAI’s \texttt{o4-mini}, \texttt{o4}, \texttt{o3}, and \texttt{o3-mini}. For comparison, we also include a smaller model: \texttt{Mistral-Small}. All experiments are conducted on the \textit{Azure AI Foundry} platform\footnote{\url{https://azure.microsoft.com/en-us/products/ai-foundry}}.

We do not aim to benchmark all existing frontier models—for instance, we do not test the Claude family despite its strong reported performance in \citet{niklaus2025swiltra}. Instead, this experiment is intended to establish a representative baseline for large reasoning models, which serves as a point of comparison for the SFT and RL experiments using small language  models \footnote{We define small language models as models with between 1 billion and 12 billion parameters.} presented in later sections.

All models are evaluated using the same prompt described in \autoref{sec::dataset}, which directly instructs the model to produce the final translation. This prompt format has been shown to be effective in eliciting accurate outputs in the Swiss legal context.

\subsection{Experiment II: Reasoning via SFT}

Experiment II investigates whether supervised fine-tuning of small language models with intermediate reasoning steps can enhance legal translation quality. To this end, we explore two SFT approaches. The first, \textbf{Simple SFT}, involves fine-tuning the SLMs solely on the correct translations. The second, \textbf{Reasoning SFT}, fine-tunes the SLMs using both the final translations and the intermediate reasoning steps that lead to them. The goal is to encourage the SLMs to \textit{think} more like a large reasoning model, potentially enabling it to produce more accurate and contextually grounded translations.

 \paragraph{Training Setup}

We fine-tune three SLMs on our dataset: \texttt{Qwen3.5 4B}, \texttt{Qwen3.5 9B}, and \texttt{Gemma 3 12B}. These models were selected for their comparable parameter scales and for representing two distinct model families.

Fine-tuning is performed using the Hugging Face \texttt{transformers}\footnote{\url{https://github.com/huggingface/transformers}} and \texttt{unsloth}\footnote{\url{https://github.com/unslothai/unsloth}} libraries, with 4-bit quantization and 8-bit AdamW optimization~\citep{loshchilov2017decoupled, dettmers20218}. Training is conducted on two NVIDIA A40 48GB GPUs. We employ Low-Rank Adaptation (LoRA)~\citep{hu2022lora, kalajdzievski2023rank} with a rank of 16 and an alpha value of 32.

\paragraph{Instruction Format}

We adopt the Alpaca-style instruction tuning format, which has been shown to improve generalization to new tasks~\citep{alpaca}. The instruction template is presented in \autoref{sec:instruc:sft}.

The key distinction in the Reasoning SFT setting lies in incorporating intermediate reasoning steps generated by DeepSeek-R1 into the model's training responses. Specifically, we extract the content enclosed between the \texttt{<think>} and \texttt{</think>} tags and pair it with the corresponding \textit{ground truth} target sentence as the final translation. This setup aims to expose the model to structured reasoning patterns that precede accurate translations.

We train the models using a maximum sequence length of 512 tokens for Simple SFT, which covers approximately 99\% of the sentences, and 1024 tokens for Reasoning SFT, which captures over 95\% of the training set (with the remainder truncated). We deliberately avoid using the maximum allowable sequence length—particularly for Reasoning SFT—as we observed that very long inputs can cause models such as those in the Qwen family to produce repetitive outputs. Limiting the input length also encourages the models to generate more concise and focused reasoning steps.

We adopt a training setup similar to that of \citet{niklaus2025swiltra}. All models are trained on the full training set to maximize data coverage. We apply sequence packing, use a weight decay of 0.01, and set the batch size between 40 and 64 depending on model capacity. Training is conducted for 10 epochs with early stopping based on validation performance (patience set to 3 epochs). We use a cosine learning rate schedule~\cite{cos-lr-scheduler} with a warmup ratio of 0.1 and a base learning rate of $1\mathrm{e}{-4}$, which we manually tune within the range of [$2\mathrm{e}{-5}$, $1\mathrm{e}{-4}$].

\paragraph{Generation}
We use \texttt{vLLM}\footnote{\url{https://github.com/vllm-project/vllm}}~\cite{vllm} for sentence generation during evaluation. After experimenting with temperature values in the range $[0, 0.7]$, we found that a temperature of 0 consistently produced the most reliable and coherent outputs across all settings. All generations were performed on an NVIDIA A40 48GB GPU. Generation over the full evaluation dataset typically takes around 30 minutes for the Simple SFT models, while the Reasoning SFT models require 3--6 hours due to longer output sequences.

\begin{table*}[ht]
\caption{Performance of frontier models for legal translation using prompting (Experiment I). }
\label{tab:results-commercial}
  \begin{center}
    \begin{tabular}[c]{lc|cccc}
      \toprule
      \multicolumn{1}{c}{\multirow{2}{*}{Model}} & Output Pricing & \multirow{2}{*}{ChrF $\uparrow$ } & \multirow{2}{*}{COMET $\uparrow$ }  & \multirow{2}{*}{METEOR $\uparrow$ } & \multirow{2}{*}{MetricX $\downarrow$ } \\
       & per 1M tokens & & & \\
      \midrule
     OpenAI o3          & \$8.00 & \textbf{64.87 $\pm$ 0.10} & \textbf{85.92 $\pm$ 0.05} & \textbf{64.82 $\pm$ 0.12}  & \textbf{2.54 $\pm$ 0.01} \\
      OpenAI 4o          & \$8.00 & 62.40 $\pm$ 0.10 & 85.35 $\pm$ 0.05 & 61.57 $\pm$ 0.13 & 2.87 $\pm$ 0.01\\
      DeepSeek-R1        & \$5.40 & 61.04 $\pm$ 0.10 & 84.35 $\pm$ 0.05 & 59.27 $\pm$ 0.12 & 3.13 $\pm$ 0.01 \\
     OpenAI o4-mini     & \$4.40  & 60.53 $\pm$ 0.10 & 84.35 $\pm$ 0.05 & 58.91 $\pm$ 0.12 & 2.84 $\pm$ 0.01 \\
   OpenAI o3-mini     & \$4.40  & 58.88 $\pm$ 0.10 & 84.21 $\pm$ 0.05 & 56.88 $\pm$ 0.12  & 3.09 $\pm$ 0.01\\
Mistral-Small-2503 & \$0.15  & 61.22 $\pm$ 0.10 & 84.41 $\pm$ 0.05 & 60.19 $\pm$ 0.12 & 3.12 $\pm$ 0.01\\

      \bottomrule
    \end{tabular}
  \end{center}
\end{table*}

\begin{figure*}[!ht]
    \centering
    \begin{tikzpicture}
        \begin{axis}[
            ybar=0pt,
            bar width=8pt,
            width=0.9\textwidth,
            height=5cm,
            enlarge x limits=0.12,
            ymin=80,
            ymax=90,
            ylabel={Mean COMET Score},
            symbolic x coords={$de \rightarrow en$, $de \rightarrow fr$, $de \rightarrow it$, $fr \rightarrow en$, $fr \rightarrow it$, $it \rightarrow en$},
            xtick=data,
            x tick label style={font=\small},
            xtick style={draw=none},
            legend style={
                at={(0.5, 1.05)},
                anchor=south, 
                legend columns=-1,
                font=\small,
                draw=none,
                /tikz/every even column/.append style={column sep=0.5cm}
            },
            legend image code/.code={
                \draw [#1] (0cm,-0.1cm) rectangle (0.6cm,0.1cm);
            },
            error bars/y dir=both,
            error bars/y explicit,
            error bars/error bar style={line width=0.5pt, black},
        ]
          \addplot[fill=plotcolour1, draw=black!30] coordinates {
              ($de \rightarrow en$, 86.09) +- (0, 0.16)
              ($de \rightarrow fr$, 83.40) +- (0, 0.10)
              ($de \rightarrow it$, 86.11) +- (0, 0.09)
              ($fr \rightarrow en$, 84.94) +- (0, 0.17)
              ($fr \rightarrow it$, 88.53) +- (0, 0.07)
              ($it \rightarrow en$, 85.51) +- (0, 0.16)
          };
          \addlegendentry{o3}

          \addplot[fill=plotcolour2, draw=black!30] coordinates {
              ($de \rightarrow en$, 85.69) +- (0, 0.15)
              ($de \rightarrow fr$, 82.79) +- (0, 0.10)
              ($de \rightarrow it$, 85.20) +- (0, 0.09)
              ($fr \rightarrow en$, 84.44) +- (0, 0.16)
              ($fr \rightarrow it$, 88.32) +- (0, 0.07)
              ($it \rightarrow en$, 84.86) +- (0, 0.16)
          };
          \addlegendentry{4o}

          \addplot[fill=plotcolour3, draw=black!30] coordinates {
              ($de \rightarrow en$, 84.89) +- (0, 0.18)
              ($de \rightarrow fr$, 81.55) +- (0, 0.11)
              ($de \rightarrow it$, 84.23) +- (0, 0.10)
              ($fr \rightarrow en$, 83.93) +- (0, 0.17)
              ($fr \rightarrow it$, 87.26) +- (0, 0.08)
              ($it \rightarrow en$, 84.28) +- (0, 0.18)
          };
          \addlegendentry{DeepSeek-R1}

          \addplot[fill=plotcolour4, draw=black!30] coordinates {
              ($de \rightarrow en$, 85.75) +- (0, 0.16)
              ($de \rightarrow fr$, 81.98) +- (0, 0.11)
              ($de \rightarrow it$, 84.20) +- (0, 0.10)
              ($fr \rightarrow en$, 84.42) +- (0, 0.17)
              ($fr \rightarrow it$, 86.38) +- (0, 0.09)
              ($it \rightarrow en$, 84.94) +- (0, 0.16)
          };
          \addlegendentry{o4-mini}

          \addplot[fill=plotcolour5, draw=black!30] coordinates {
              ($de \rightarrow en$, 85.27) +- (0, 0.16)
              ($de \rightarrow fr$, 81.35) +- (0, 0.10)
              ($de \rightarrow it$, 83.92) +- (0, 0.09)
              ($fr \rightarrow en$, 84.08) +- (0, 0.17)
              ($fr \rightarrow it$, 87.11) +- (0, 0.08)
              ($it \rightarrow en$, 84.28) +- (0, 0.16)
          };
          \addlegendentry{o3-mini}

          \addplot[fill=plotcolour6, draw=black!30] coordinates {
              ($de \rightarrow en$, 84.91) +- (0, 0.18)
              ($de \rightarrow fr$, 81.76) +- (0, 0.11)
              ($de \rightarrow it$, 84.31) +- (0, 0.09)
              ($fr \rightarrow en$, 83.72) +- (0, 0.17)
              ($fr \rightarrow it$, 87.30) +- (0, 0.08)
              ($it \rightarrow en$, 83.94) +- (0, 0.17)
          };
          \addlegendentry{Mistral-Small-2503}
        \end{axis}
    \end{tikzpicture}
    \caption{Comparison of translation quality (mean of COMET score) across language pairs for different frontier models evaluated on the test dataset.}
    \label{fig:results_expeI}
\end{figure*}
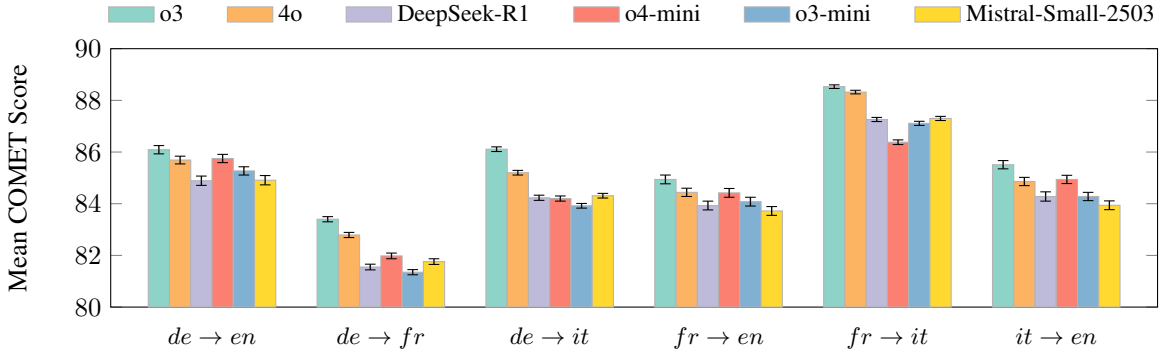

\subsection{Experiment III: Reasoning via Reinforcement Learning}

We use the same training and validation datasets as in Experiment II.

The reinforcement learning employs GRPO with simple verifiable rewards in order to promote the reasoning alongside
a high quality translation. Similar to DeepSeek-R1, two rule-based rewards have been used:

\begin{itemize}
  \item \textbf{Format:} To make the translation easily extractable while also enforcing the presence of the reasoning,
    the model must put its thinking process inside the \texttt{<think></think>} tag and its final translation inside the
    \texttt{<translation></translation>} tag. The presence of each tag gives a reward of 0.5, with an additional 0.5 if they occur
    in the correct order. This is to ensure that the thinking is done before given the final translation, such that it
    can potentially improve the translation rather than given an explanation after the fact.
  \item \textbf{Translation quality:} A reward for the quality of the translation is paramount to guide the model to
    produce better responses. The final translation is extracted from the response, which is then compared to
    the ground truth. We chose the ChrF score to assess the quality of the translation and used it directly as a reward,
    since the values are in the range $[0, 1]$. This results in a continuous scale for the reward, which gives extra
    nuance to the differences in responses, allowing a more fine-grained adaptation.
\end{itemize}

A group size of $G = 4$ is used, meaning that for each input sentence four different responses are sampled which are
then compared to each other based on the rewards. Due to the increased computational demands for these generations,
only a single epoch was performed compared to the 10 epochs for the SFT.
The training parameters remain unchanged, with the exception of the learning
rate, which needed to be reduced to $2\mathrm{e}{-5}$ as the higher learning rate led to
training instabilities.

\section{Results}

\paragraph{Performance of State-of-the-Art Reasoning Models}

We report the performance of large reasoning models in \autoref{tab:results-commercial}. As the exact model sizes are not publicly disclosed, we use the output cost per million tokens as a proxy measure\footnote{Experiments and pricing estimates correspond to July 2025.}. Among the evaluated models, the \texttt{o3} model achieves the highest performance across all metrics but is also the most expensive. In contrast, the \texttt{Mistral-Small-2503} model is approximately ten times cheaper while delivering comparable results, making it the most cost-effective option.

When examining performance across language pairs (\autoref{fig:results_expeI}), we find that the relative ranking of models remains largely consistent across pairs. All models achieve their best results on the $fr \rightarrow it$ pairs, likely due to the close linguistic relationship between these two Romance languages. The \texttt{o4-mini} model shows particularly strong performance when translating into English, whereas the \texttt{mistral-small-2503} model performs better on translations into French or Italian.

\begin{table*}[ht]
\caption{Comparison of performance of SFT and RL in legal translation (Experiments II and III).}
\label{tab:results-finetune}
  \begin{center}
    \begin{tabular}[c]{lc|cccc}
      \toprule
      \multicolumn{1}{c}{Model} & Method & ChrF $\uparrow$ & COMET $\uparrow$ & METEOR $\uparrow$ & MetricX $\downarrow$  \\
      \midrule
            \multirow{4}{*}{Qwen3.5 4B}
   & Base Model     & 50.53 $\pm$ 0.18 & 77.05 $\pm$ 0.15 & 47.21 $\pm$ 0.19 & 5.13 $\pm$ 0.04 \\
    & Simple SFT     & 50.03 $\pm$ 0.12 & 76.47 $\pm$ 0.09 & 47.96 $\pm$ 0.13 & 6.07 $\pm$ 0.03 \\
     & Reasoning SFT  & 42.84 $\pm$ 0.16 & 69.88 $\pm$ 0.14 & 39.99 $\pm$ 0.16 & 7.56 $\pm$ 0.03 \\
     & RL             & 54.64 $\pm$ 0.10 & 79.47 $\pm$ 0.07 & 50.98 $\pm$ 0.12 & 4.95 $\pm$ 0.02 \\
    
    \hline
      \multirow{4}{*}{Qwen3.5 9B} & Base Model  & 51.76 $\pm$ 0.19 & 77.99 $\pm$ 0.16 & 49.08 $\pm$ 0.20 & 4.66 $\pm$ 0.04 \\
                                    & Simple SFT & 51.38 $\pm$ 0.12 & 77.25 $\pm$ 0.10 & 49.36 $\pm$ 0.14 & 5.61 $\pm$ 0.03 \\
                                     & Reasoning SFT    & 49.42 $\pm$ 0.12 & 76.88 $\pm$ 0.10 & 47.59 $\pm$ 0.14 & 5.69 $\pm$ 0.03 \\
                                    & RL               & 56.02 $\pm$ 0.10 & 81.52 $\pm$ 0.06 & 53.66 $\pm$ 0.12 & 4.07 $\pm$ 0.02 \\
      \hline
      \multirow{4}{*}{Gemma 3 12B} & Base Model         & 56.73 $\pm$ 0.10 & 82.49 $\pm$ 0.05 & 53.20 $\pm$ 0.12  & 3.70 $\pm$ 0.02 \\
                                  & Simple SFT         & 55.86 $\pm$ 0.11 & 82.45 $\pm$ 0.06 & 54.29 $\pm$ 0.12 & 3.71 $\pm$ 0.02 \\
                                  & Reasoning SFT      & 53.29 $\pm$ 0.14 & 79.02 $\pm$ 0.12 & 51.74 $\pm$ 0.16  & 4.53 $\pm$ 0.03 \\
                                  & RL                 & \textbf{57.32 $\pm$ 0.10} & \textbf{83.08 $\pm$ 0.05} & \textbf{54.92 $\pm$ 0.12} & \textbf{3.32 $\pm$ 0.01} \\

      \bottomrule
    \end{tabular}
  \end{center}
\end{table*}

\begin{figure*}[ht]
    \centering
    \pgfplotslegendfromname{sharedlegend}
    \begin{tikzpicture}
        \begin{axis}[
            name=ax1,
            title={\textbf{Qwen3.5 9B}},
            title style={yshift=-4pt},
            ybar=0pt,
            bar width=6pt, 
            width=0.48\textwidth,
            height=5cm,
            enlarge x limits=0.15,
            ymin=70,
            ymax=90, 
            ylabel={Mean COMET Score},
            symbolic x coords={$de \rightarrow en$, $de \rightarrow fr$, $de \rightarrow it$, $fr \rightarrow en$, $fr \rightarrow it$, $it \rightarrow en$},
            xtick=data,
            x tick label style={font=\small, rotate=45, anchor=north east},
            xtick style={draw=none},
            legend to name=sharedlegend,
            legend style={
                legend columns=-1,
                font=\small,
                draw=none,
                /tikz/every even column/.append style={column sep=0.5cm}
            },
            legend image code/.code={
                \draw [#1] (0cm,-0.1cm) rectangle (0.6cm,0.1cm);
            },
            error bars/y dir=both,
            error bars/y explicit,
            error bars/error bar style={line width=0.5pt, black},
        ]
            \addplot[fill=plotcolour1, draw=black!30] coordinates {
                ($de \rightarrow en$, 82.69) +- (0, 0.41)
                ($de \rightarrow fr$, 73.85) +- (0, 0.33)
                ($de \rightarrow it$, 76.37) +- (0, 0.33)
                ($fr \rightarrow en$, 80.42) +- (0, 0.49)
                ($fr \rightarrow it$, 80.52) +- (0, 0.30)
                ($it \rightarrow en$, 81.85) +- (0, 0.42)
            };
            \addlegendentry{Base}

            \addplot[fill=plotcolour2, draw=black!30] coordinates {
                ($de \rightarrow en$, 81.54) +- (0, 0.31)
                ($de \rightarrow fr$, 74.06) +- (0, 0.18)
                ($de \rightarrow it$, 75.18) +- (0, 0.20)
                ($fr \rightarrow en$, 81.23) +- (0, 0.29)
                ($fr \rightarrow it$, 79.82) +- (0, 0.19)
                ($it \rightarrow en$, 80.64) +- (0, 0.33)
            };
            \addlegendentry{Simple SFT}

            \addplot[fill=plotcolour3, draw=black!30] coordinates {
                ($de \rightarrow en$, 80.34) +- (0, 0.33)
                ($de \rightarrow fr$, 73.27) +- (0, 0.19)
                ($de \rightarrow it$, 74.73) +- (0, 0.21)
                ($fr \rightarrow en$, 80.15) +- (0, 0.34)
                ($fr \rightarrow it$, 80.42) +- (0, 0.19)
                ($it \rightarrow en$, 80.01) +- (0, 0.34)
            };
            \addlegendentry{Reasoning SFT}

            \addplot[fill=plotcolour4, draw=black!30] coordinates {
                ($de \rightarrow en$, 84.86) +- (0, 0.17)
                ($de \rightarrow fr$, 78.18) +- (0, 0.12)
                ($de \rightarrow it$, 80.93) +- (0, 0.12)
                ($fr \rightarrow en$, 83.45) +- (0, 0.19)
                ($fr \rightarrow it$, 83.72) +- (0, 0.11)
                ($it \rightarrow en$, 83.71) +- (0, 0.18)
            };
            \addlegendentry{RL}
        \end{axis}

        \begin{axis}[
            name=ax2,
            at={(ax1.outer east)},
            anchor=outer west,
            title={\textbf{Gemma 3 12B}},
            title style={yshift=-4pt},
            ybar=0pt,
            bar width=6pt, 
            width=0.48\textwidth,
            height=5cm,
            enlarge x limits=0.15,
            ymin=70,
            ymax=90, 
            symbolic x coords={$de \rightarrow en$, $de \rightarrow fr$, $de \rightarrow it$, $fr \rightarrow en$, $fr \rightarrow it$, $it \rightarrow en$},
            xtick=data,
            x tick label style={font=\small, rotate=45, anchor=north east},
            xtick style={draw=none},
            error bars/y dir=both,
            error bars/y explicit,
            error bars/error bar style={line width=0.5pt, black},
        ]
            \addplot[fill=plotcolour1, draw=black!30] coordinates {
                ($de \rightarrow en$, 84.95) +- (0, 0.00)
                ($de \rightarrow fr$, 79.43) +- (0, 0.00)
                ($de \rightarrow it$, 82.38) +- (0, 0.00)
                ($fr \rightarrow en$, 83.74) +- (0, 0.00)
                ($fr \rightarrow it$, 84.40) +- (0, 0.00)
                ($it \rightarrow en$, 84.21) +- (0, 0.00)
            };

            \addplot[fill=plotcolour2, draw=black!30] coordinates {
                ($de \rightarrow en$, 84.49) +- (0, 0.19)
                ($de \rightarrow fr$, 78.81) +- (0, 0.12)
                ($de \rightarrow it$, 82.42) +- (0, 0.10)
                ($fr \rightarrow en$, 83.75) +- (0, 0.19)
                ($fr \rightarrow it$, 84.98) +- (0, 0.09)
                ($it \rightarrow en$, 83.90) +- (0, 0.19)
            };

            \addplot[fill=plotcolour3, draw=black!30] coordinates {
                ($de \rightarrow en$, 84.31) +- (0, 0.00)
                ($de \rightarrow fr$, 79.79) +- (0, 0.00)
                ($de \rightarrow it$, 81.99) +- (0, 0.16)
                ($fr \rightarrow en$, 74.39) +- (0, 0.00)
                ($fr \rightarrow it$, 74.00) +- (0, 0.00)
                ($it \rightarrow en$, 82.27) +- (0, 0.29)
            };

            \addplot[fill=plotcolour4, draw=black!30] coordinates {
                ($de \rightarrow en$, 85.30) +- (0, 0.16)
                ($de \rightarrow fr$, 79.79) +- (0, 0.11)
                ($de \rightarrow it$, 83.04) +- (0, 0.10)
                ($fr \rightarrow en$, 83.89) +- (0, 0.17)
                ($fr \rightarrow it$, 85.46) +- (0, 0.09)
                ($it \rightarrow en$, 84.03) +- (0, 0.18)
            };

        \end{axis}
    \end{tikzpicture}
    \caption{Comparison of translation quality (mean of COMET score) for Qwen3.5 9B (left) and Gemma 3 12B (right) across different language pairs and training procedures.}
    \label{fig:combined_models_lang}
\end{figure*}
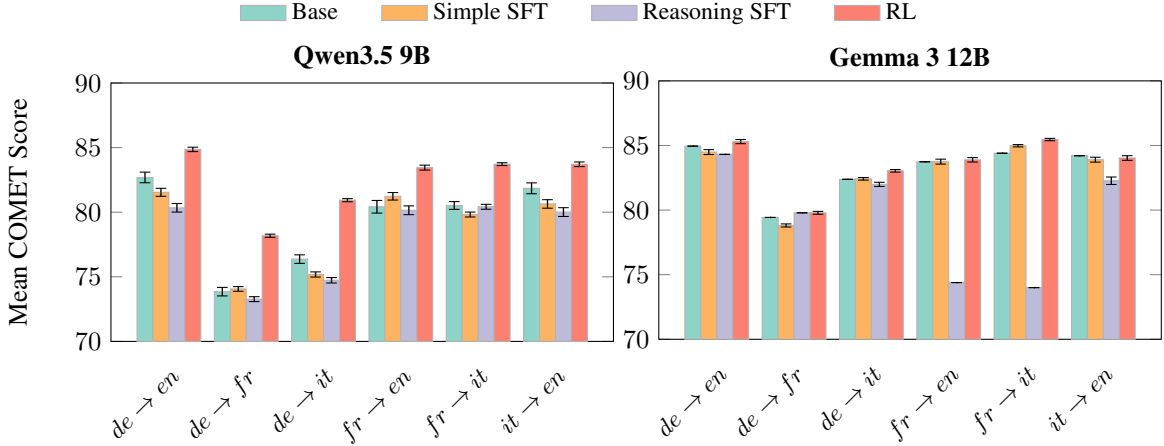

\paragraph{Performance of SFT vs. RL}

\autoref{tab:results-finetune} reports the overall results for SFT and RL using four evaluation metrics: ChrF, COMET, METEOR and MetricX. We find that RL consistently improves translation quality across all models and achieves the best overall performance. In particular, RL with Gemma 3 12B delivers the strongest results among the smaller models, performing just below the frontier model \texttt{o4-mini}. This demonstrates that reasoning-augmented training can substantially narrow the performance gap between small models and state-of-the-art LLMs.

When comparing RL to SFT, we observe that RL consistently outperforms both SFT variants across all models.
Examining the two SFT approaches more closely, we find that Reasoning SFT degrades the quality of the translation in comparison to Simple SFT. This is especially noticeable on the smallest model \texttt{Qwen3.5 4B}, where the gap of every metric is much larger than in the larger models.
Although Simple SFT outperforms the Reasoning SFT, it performs marginally worse than the base model. This means that it is more beneficial to use the pretrained models unaltered than training them with SFT.

In terms of performance differences across language pairs, as illustrated in
\autoref{fig:combined_models_lang} for Qwen3.5 9B and Gemma 3 12B, we observe
that RL performs on par with or slightly better than the baselines for most
language pairs. For Qwen3.5 9B, all language pairs exhibit a consistent pattern:
RL achieves the best performance, followed by Simple SFT and the base model
which outperform each other in different categories, and lastly Reasoning SFT.
For Gemma 3 12B, the base model is much closer to RL, and technically surpasses
it for $it \rightarrow en$ by the tiniest margin, although that would be
considered to be on par. The strongest overall results for Qwen3.5 9B and Gemma
3 12B can be observed in the language pairs $de \rightarrow en$, whereas the 
$fr \rightarrow it$ pairs exhibit the largest difference compared to the frontier models, which are their best performing language pairs.

\paragraph{Comparison Across Metrics}

Comparing the different evaluation metrics, we observe that while each captures a different aspect of translation quality—lexical overlap (ChrF), semantic similarity (COMET / MetricX), and linguistic fluency (METEOR)—the overall ranking of methods remains consistent across all metrics. Although absolute scores vary, RL consistently outperforms other approaches. This trend aligns with the results from Experiment~I, where the strongest models also achieve top performance across all evaluation criteria.

\paragraph{Human Evaluation}

\begin{table}[!ht]
\centering
\caption{Pearson correlation between human evaluation scores and automatic evaluation metrics. COMET shows the strongest correlation with expert judgments on the legal translation dataset.}
\small
\begin{tabular}{lcc}
\toprule
\textbf{Metric} & \textbf{Pearson Correlation ($r$)} & \textbf{p-value} \\
\midrule
ChrF     & 0.05  & 0.54 \\
METEOR   & 0.07  & 0.40 \\
COMET    & 0.24  & 0.00 \\
MetricX  & -0.13 & 0.11 \\
\bottomrule
\end{tabular}
\label{tab:human_correlation}
\end{table}

We sampled 240 translation pairs covering all language combinations present in the dataset. A legal expert manually assessed the translation quality on a scale from 1 to 10, considering factors such as accuracy, fluency, and preservation of legal meaning. We then computed the Pearson correlation between the human annotation scores and the automatic evaluation metrics to assess their alignment with expert judgments. As shown in \autoref{tab:human_correlation}, COMET achieves the highest correlation with human evaluations, indicating a stronger alignment with expert judgments on the legal translation dataset.

\section{Discussions}

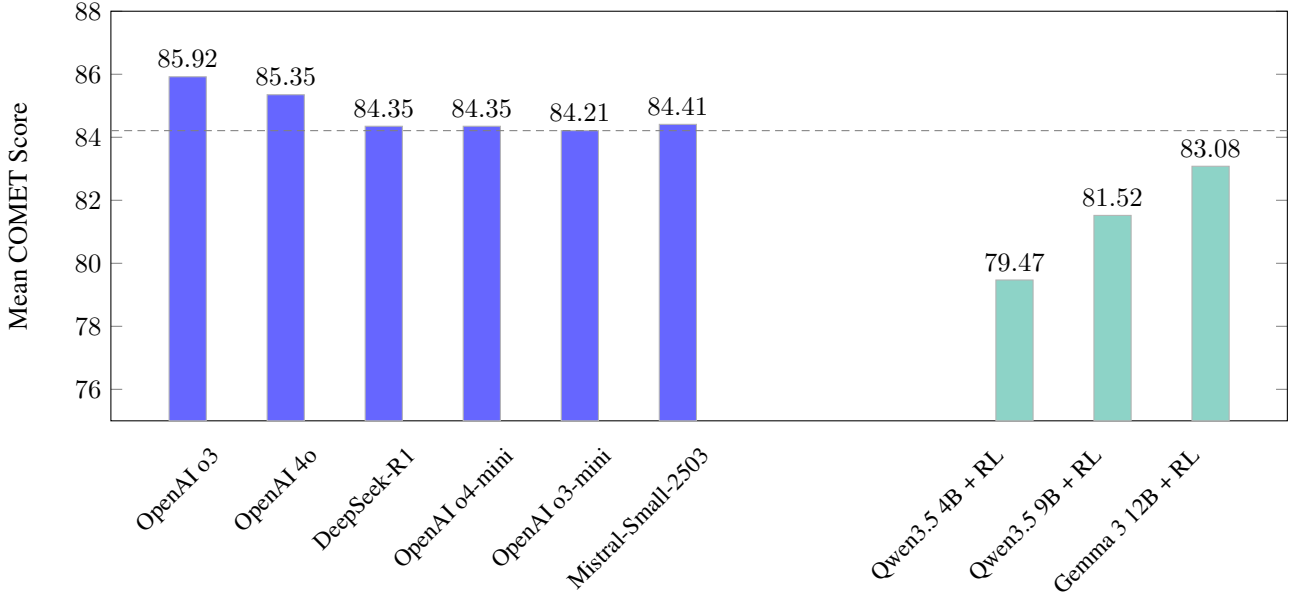
\begin{figure*}[h!t]
    \centering
    \begin{tikzpicture}
        \begin{axis}[
            ybar,
            bar width=14pt,
            width=\textwidth,
            height=7cm,
            ymin=75,
            ymax=88,
            ylabel={Mean COMET Score},
            xmin=0, 
            xmax=12,
            xtick={1, 2, 3, 4, 5, 6,  9, 10, 11},
            xticklabels={
                OpenAI o3,
                OpenAI 4o,
                DeepSeek-R1,
                OpenAI o4-mini,
                OpenAI o3-mini,
                Mistral-Small-2503,
                Qwen3.5 4B + RL,
                Qwen3.5 9B + RL,
                Gemma 3 12B + RL
            },
            x tick label style={
                rotate=45,
                anchor=north east,
                font=\small
            },
            xtick style={draw=none},
            nodes near coords,
            nodes near coords align={vertical},
        ]

        \addplot[
            fill=blue!60,
            draw=black!30
        ] coordinates {
            (1, 85.92)
            (2, 85.35)
            (3, 84.35)
            (4, 84.35)
            (5, 84.21)
            (6, 84.41)
        };

        \addplot[
            fill=plotcolour1,
            draw=black!30
        ] coordinates {
            (9, 79.47)
            (10, 81.52)
            (11, 83.08)
        };

        \draw[densely dashed, black!50] (axis cs:0, 84.21) -- (axis cs:12, 84.21);

        \end{axis}
    \end{tikzpicture}
    \caption{Performance comparison of RL enhanced small models with frontier models, using the COMET score as a benchmark.}
    \label{fig:cost-quality}
\end{figure*}

\subsection{Leveraging Reasoning in Translation}

Translation often requires careful refinement and review to ensure alignment with normative language \cite{ordudari2007translation,nida1964toward}. This is particularly important in technical domains such as the legal field, where precision in terminology and strict adherence to legal conventions are essential. These requirements extend beyond surface-level semantic equivalence. This motivates the integration of reasoning into the translation process, as prior studies have shown that adding self-reaffine improve translation \cite{chen2023iterative,wang2024drt}.

In our training dataset for the Reasoning SFT setup, we examined the reasoning steps generated by DeepSeek. We found that the model produces coherent, multi-step reasoning sequences that often resemble the workflow of a human translator. These typically include decomposing the sentence, translating individual components, recomposing the full output, and conducting a final review and refinement.

While the objective of Reasoning SFT is to encourage a more structured and deliberate translation process, our results indicate that fine-tuning with explicit reasoning steps degrades the performance rather than improving it. These models may already possess strong intrinsic reasoning abilities that were obtained during their pre-training, therefore, the addition of shallow or formulaic reasoning steps offers little added value.

Another concern is that current reasoning models, while demonstrating emergent reasoning abilities \cite{wei2022emergent, webb2023emergent}, often fail to provide genuine explanatory depth. In particular, these capacities are often considered superficial, and models are unlikely to generate ``how-actually'' explanations that reflect true causal or conceptual understanding \cite{musker2024llms}. Providing explicit reasoning steps during training or prompting does not necessarily incentivize models to engage in authentic reasoning. Instead, they may simply learn to reproduce reasoning-like patterns without acquiring the underlying cognitive mechanisms  such steps are meant to encourage.

\subsection{Reinforcement Learning vs. Supervised Fine-Tuning}

We observe that reinforcement learning consistently outperforms supervised fine-tuning in our experiments. This suggests that RL may provide a more effective mechanism for enabling models to internalize reasoning strategies through the reward optimization process.

In the broader literature, the differences between SFT and RL have been explored across a range of tasks, with RL predominantly showing improved generalization capabilities. \citet{sft-vs-rl} compared them in an arithmetic reasoning card game and a real-world navigation scenario, which require text-based and visual reasoning, and came to the conclusion that SFT tends to memorize the training data and therefore struggles to generalize to out-of-distribution data, while RL exhibits more generalized reasoning capabilities.
Similarly, \citet{rule-based-rl-document-classification} showed that RL generalizes better to previously unseen classes and out-of-distribution samples on the task of document classification.
In the medical domain, Med-RLVR~\cite{med-rlvr} observed a similar trend for medical multiple-choice questions where RL improves the accuracy on out-of-distribution data by 8 points compared to SFT, indicating that it also applies to domain specific language.
For general machine translation, MT-R1-Zero~\cite{mt-r1-zero} successfully applied RL with competitive performance to other models trained with SFT, while observing stronger generalization to out-of-distribution tasks, including languages the model was not trained on.

\subsection{Global Quality and Cost-Quality Comparisons}

\paragraph{Pure Quality}

Figure \ref{fig:cost-quality} shows that commercial frontier models reach COMET scores in the 84.21 to 85.94 range, whereas small models enhanced with the reinforcement learning method fall in the 79.47 to 83.08 range. There is nearly an overlap, as the gap keeps narrowing. 

Considering quality exclusively (that is, not considering cost of model for the moment), performance of small enhanced models could potentially be raised by fine-tuning parameters such as the value of the reward, the group size, etc. Hence, enhanced small models might be competitive for NMT. 

It might be tempting to try to fill the performance gap by using slightly larger RL-enhanced models. However, referring to Table \ref{tab:results-finetune}, one can remark that the relative enhancement provided by RL shows diminishing returns as the language model gets larger. The quality of Qwen3.5 4B evolves from 77.05 to 79.47 (\textit{+2.47}) in terms of COMET score, while Gemma3 12B only improves by \textit{+0.59} (from 82.49 to 83.08). We plan to investigate this further. 

\paragraph{Cost-Quality Trade-off} In contrast to commercial models, whose usage costs are known (see Table \ref{tab:results-commercial}), the costs associated with the enhanced models evaluated in this study remain unknown. Nevertheless, assuming that reinforcement learning (RL) enhancement has only a marginal impact on computational complexity, these costs are expected to represent only a fraction of those incurred by frontier models comprising several hundred billion parameters. Therefore, once the focus shifts from a quality-at-any-cost perspective to a cost–quality trade-off, our results indicate that RL-enhanced models emerge as a compelling and viable design option.

\section{Conclusion}

In this work, we investigate multiple strategies of leveraging reasoning to enhance translation quality, with a specific focus on the Swiss legal domain—a high-stakes, linguistically complex setting for NMT systems. We design and evaluate three complementary approaches to reasoning integration: prompting frontier models, SFT with intermediate reasoning steps, and RL using reward-based feedback.

Our results show that although frontier models still achieve the strongest overall translation quality, reasoning-augmented training methods—especially RL—substantially narrow the performance gap. Across all evaluated model architectures, RL consistently surpasses SFT, demonstrating that reinforcement learning with verifiable rewards is a highly effective approach for improving legal translation quality and enabling smaller open-source models to acquire stronger reasoning capabilities. Notably, our best-performing open-source models use more than 100× fewer parameters than frontier models while achieving competitive translation quality. These findings provide strong empirical evidence that explicit reasoning and RL-based optimization can significantly advance domain-specific neural machine translation.

Future work will focus on improving reinforcement learning methods for legal machine translation, as well as optimizing the trade-off between training and inference costs and translation quality. This includes investigating more effective reward designs, more efficient training strategies, and approaches that improve translation quality while reducing computational requirements.

\section*{Limitations}

We rely on four cost-efficient evaluation metrics to assess translation quality: ChrF, METEOR, COMET, and MetricX. Although more fine-grained evaluation methods—such as using LLMs as judges—may provide deeper insights, they are resource-intensive and less scalable. Additionally, we fine-tune only a handful of commonly used models. Future work could explore models of varying sizes and architectures to gain a broader understanding of performance trends. Finally, our RL experiments use simple, rule-based rewards; more complex or adaptive reward functions may offer additional gains and warrant further investigation.

\section*{Impact Statement}

This work advances legal machine translation by demonstrating that reinforcement learning with verifiable rewards can improve translation quality beyond supervised fine-tuning. By focusing on open-source models, our approach contributes to making high-quality legal translation systems more accessible, transparent, and reproducible for researchers, institutions, and practitioners who rely on open technologies.

\section*{Acknowledgements}

The authors would like to thank the Hasler Foundation and the Mercator Foundation Switzerland for their generous financial support.

\bibliography{references}
\bibliographystyle{icml2026}

\newpage
\appendix
\onecolumn
\section{Training Set Language Pair Distribution Statistics}

\begin{table}[h]
\centering
\caption{Statistics of language pair distribution in the training set used for the experiments.}
\begin{tabular}{l r}
\toprule
\textbf{Language Pair} & \textbf{Count} \\
\midrule
fr $\rightarrow$ it & 8\,152 \\
de $\rightarrow$ fr & 5\,384 \\
de $\rightarrow$ it & 4\,252 \\
de $\rightarrow$ en & 65 \\
fr $\rightarrow$ en & 65 \\
it $\rightarrow$ en & 63 \\
\hline
\textbf{Total} & \textbf{18\,981} \\
\bottomrule
\end{tabular}
\label{tab:lang-pair-counts}
\end{table}

\section{Prompt}  \label{sec:instruc:sft}

\begin{quote}
\begin{lstlisting}[breaklines=true, breakatwhitespace=true, breakindent=0pt, basicstyle=\ttfamily, columns=fullflexible]
### Instruction
Translate the following sentence from {lang_source} to {lang_target} while respecting Swiss legal parlance. Respond in the following format:
<think>
...
</think>
<translation>...</translation>

Explain your reasoning inside the <think> tag and give your final translation in the <translation> tag.

### Input
{source}
\end{lstlisting}
\end{quote}

\end{document}